\documentclass[letterpaper, 10 pt, conference]{ieeeconf}  
\usepackage{times}
\usepackage{epsfig}
\usepackage{graphicx}
\usepackage{amsmath}
\usepackage{amssymb}
\usepackage{soul}
\usepackage{bm}
\usepackage{comment}

\usepackage{algorithm}
\usepackage{algorithmic}

\usepackage{caption}
\usepackage{subcaption}

\graphicspath{ {./figs/} }

\DeclareMathOperator*{\argmax}{argmax}

\usepackage{color,soul}

\IEEEoverridecommandlockouts                              

\title{\LARGE \bf
Asynchronous Optimisation for Event-based Visual Odometry
}

\author{Daqi Liu$^{\dagger}$, Alvaro Parra$^{\dagger}$, Yasir Latif$^{\dagger}$, Bo Chen$^{\dagger}$, Tat-Jun Chin$^{\dagger,\ddagger}$, Ian Reid$^{\dagger}$
\thanks{$^{\dagger}$School of Computer Science, The University of Adelaide.}%
\thanks{$^{\ddagger}$SmartSat CRC Professorial Chair of Sentient Satellites.}%
}

\newcommand{\cK}{\mathcal{K}}
\newcommand{\cP}{\mathcal{P}}

\newcommand{\GP}{\mathcal{GP}}

\DeclareMathAlphabet\mathbfcal{OMS}{cmsy}{b}{n}

\begin{document}

\maketitle
\thispagestyle{empty}
\pagestyle{empty}

\begin{abstract}
Event cameras open up new possibilities for robotic perception due to their low latency and high dynamic range. On the other hand, developing effective event-based vision algorithms that fully exploit the beneficial properties of event cameras remains work in progress. In this paper, we focus on event-based visual odometry (VO). While existing event-driven VO pipelines have adopted continuous-time representations to asynchronously process event data, they either assume a known map, restrict the camera to planar trajectories, or integrate other sensors into the system. Towards map-free event-only monocular VO in $SE(3)$, we propose an asynchronous structure-from-motion optimisation back-end. Our formulation is underpinned by a principled joint optimisation problem involving \emph{non-parametric} Gaussian Process motion modelling and incremental \emph{maximum a posteriori} inference. A high-performance incremental computation engine is employed to reason about the camera trajectory with every incoming event. We demonstrate the robustness of our asynchronous back-end in comparison to frame-based methods which depend on accurate temporal accumulation of measurements.

\end{abstract}

\section{Introduction}\label{sec:intro}

Many robotic perception capabilities are developed for regular cameras, which are affected by blurring or lack of contrast due to high-speed manoeuvres or low-light settings. By asynchronously detecting intensity changes, event cameras have low latency and high dynamic range, which can help mitigate the above issues. However, the very different sensing principle calls for novel processing methods~\cite{gallego20}.

In this paper, we focus on event-based monocular visual odometry (VO). An event camera outputs a time-continuous stream of events $\mathcal{E} = \{e_1, e_2, \dots \}$, where each event 
\begin{align}
    e_k = (z_k, t_k, p_k)
\end{align}
is an observation of a change in intensity at pixel $z_k \in \mathbb{R}^2$ at time $t_k \in \mathbb{R}_{+}$ with polarity $p_k \in \{-1, +1\}$. The goal of VO is to recover the motion of the camera that gave rise to~$\mathcal{E}$.

\begin{figure} [h]
\centering
 \begin{subfigure}[b]{\linewidth}
 \centering
 \includegraphics[width=\textwidth]{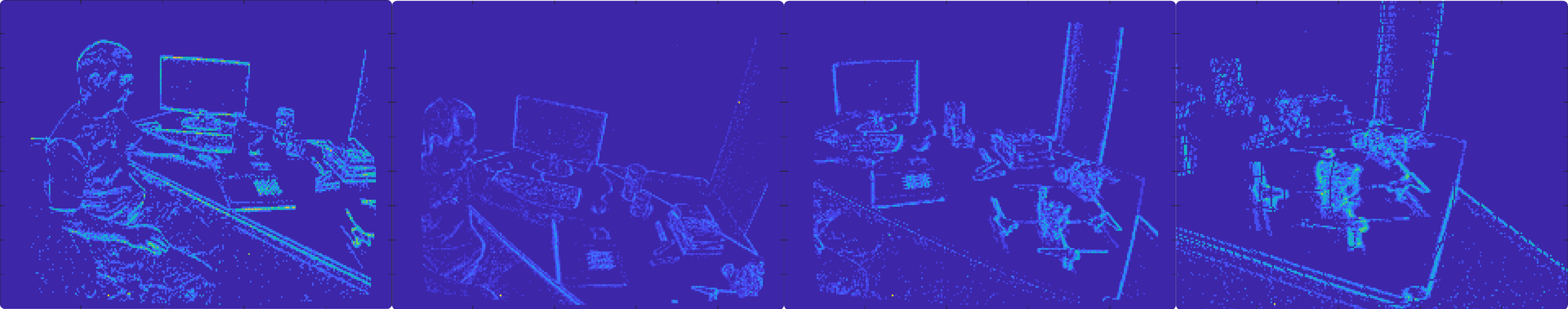}
 \caption{Event images to visualise the event data used~\cite{mueggler2017event}.}
 \label{fig:ci}
 \end{subfigure}
 \begin{subfigure}[b]{.47\linewidth}
 \centering
 \includegraphics[width=\textwidth]{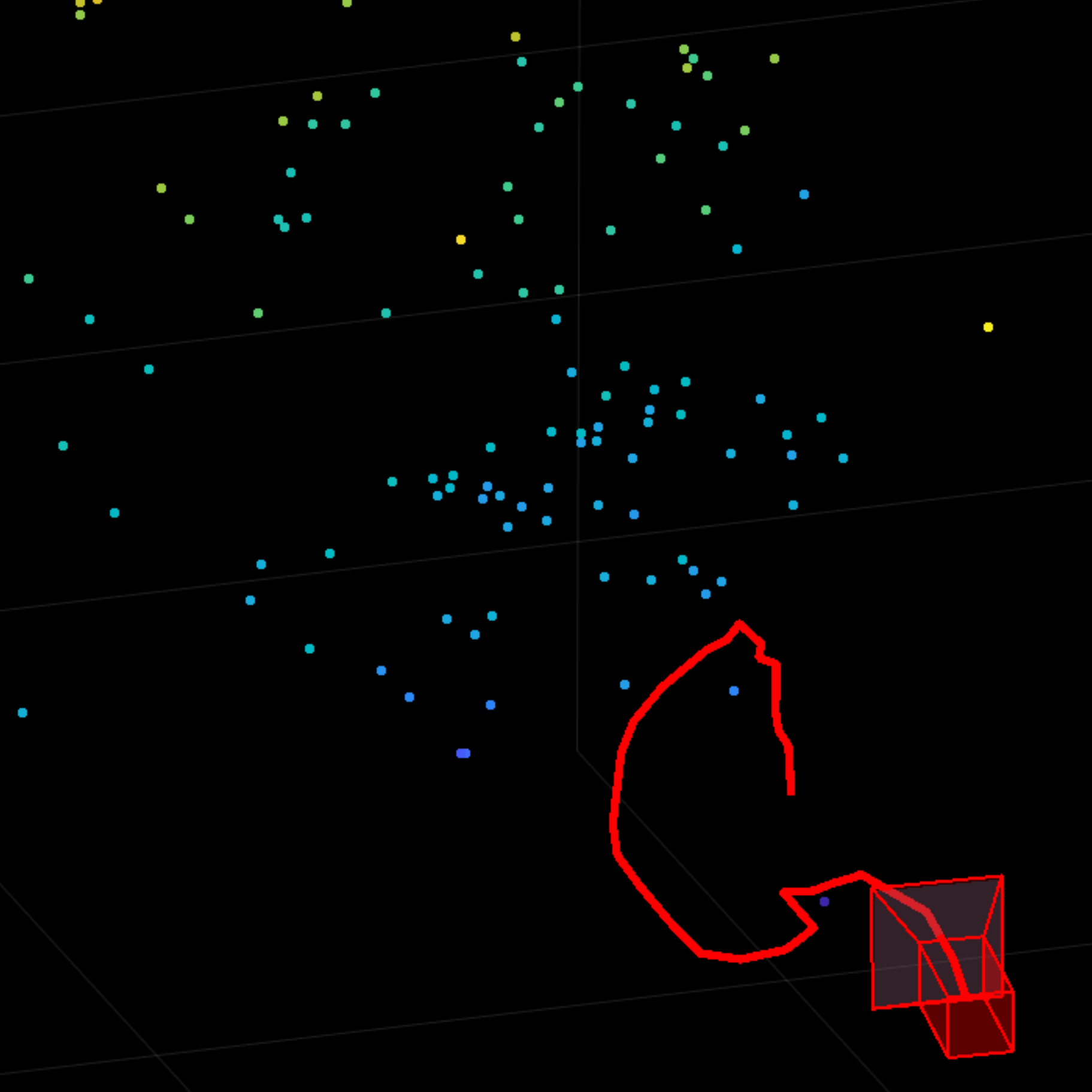}
 \caption{Frame-based method with suboptimal batching.}
 \label{fig:fb}
 \end{subfigure}
 \;\;
 \begin{subfigure}[b]{.47\linewidth}
 \centering
 \includegraphics[width=\textwidth]{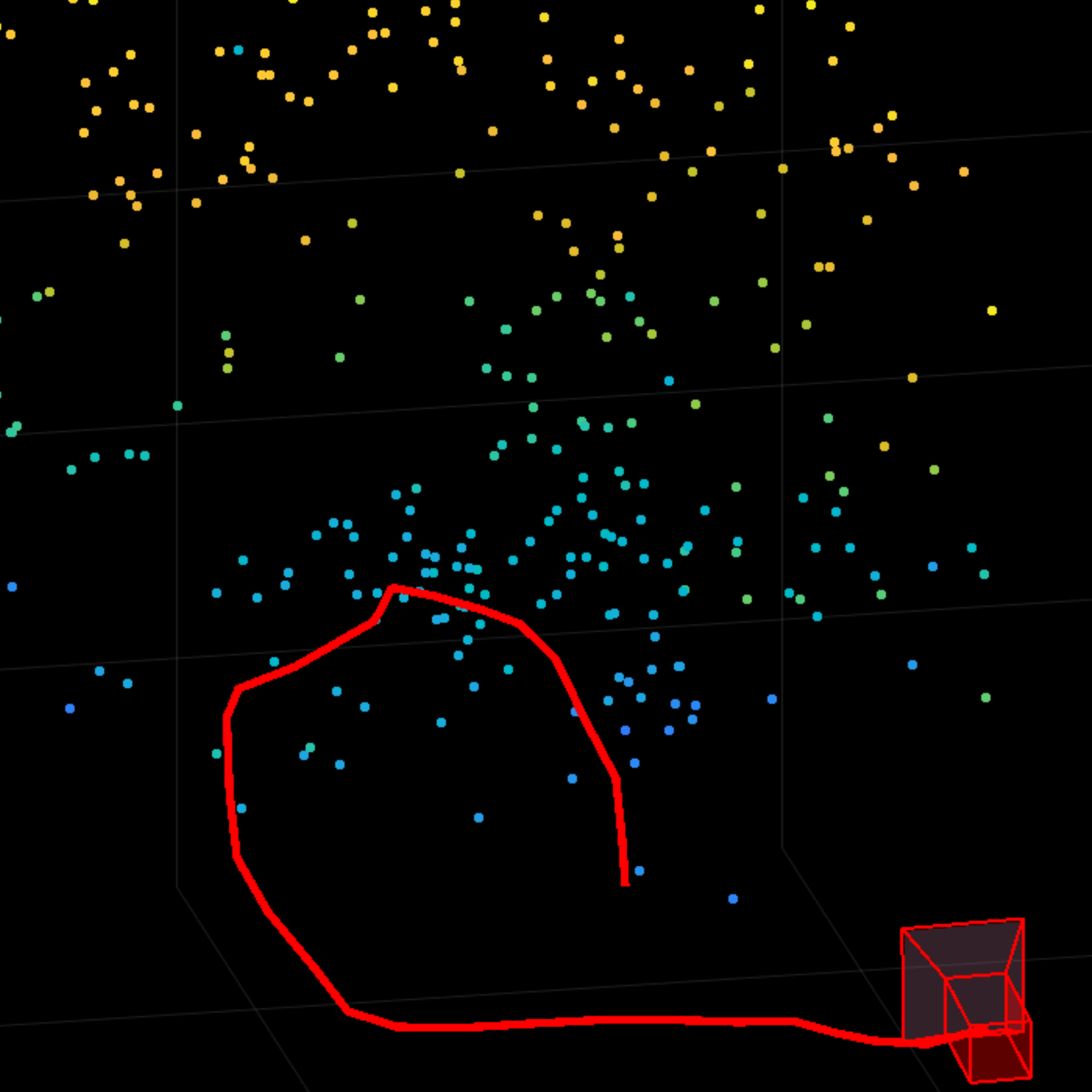}
 \caption{Asynchronous optimisation that does not require batching.}
 \label{fig:ed}
 \end{subfigure}
 \caption{(a) Input data for event-based VO. (b) The frame-based method is accurate due to suboptimal batching. Note that we are not claiming it is difficult to set the batching parameter in this specific event data; rather, we are showing the negative effects on frame-based VO \emph{given} suboptimal batching. (c) By asynchronously processing the data using our CT SfM formulation (no batching required), our event-driven approach can robustly track the camera in $SE(3)$.}
 \label{fig:eovo}
\end{figure}


\begin{figure*} [h]
\centering
 \begin{subfigure}[b]{.3\linewidth}
 \centering
 \includegraphics[width=\textwidth]{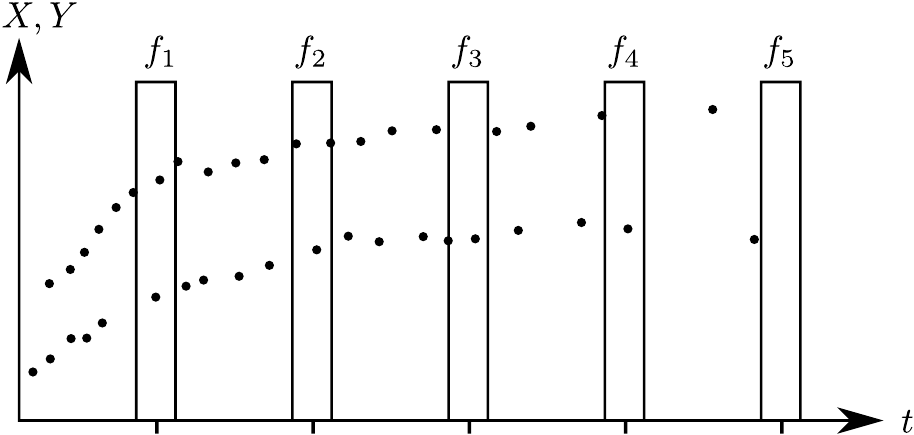}
 \caption{Batching for frame-based optimisation}
 \label{fig:fbopt}
 \end{subfigure}  \quad
 \begin{subfigure}[b]{.3\linewidth}
 \centering
 \includegraphics[width=\textwidth]{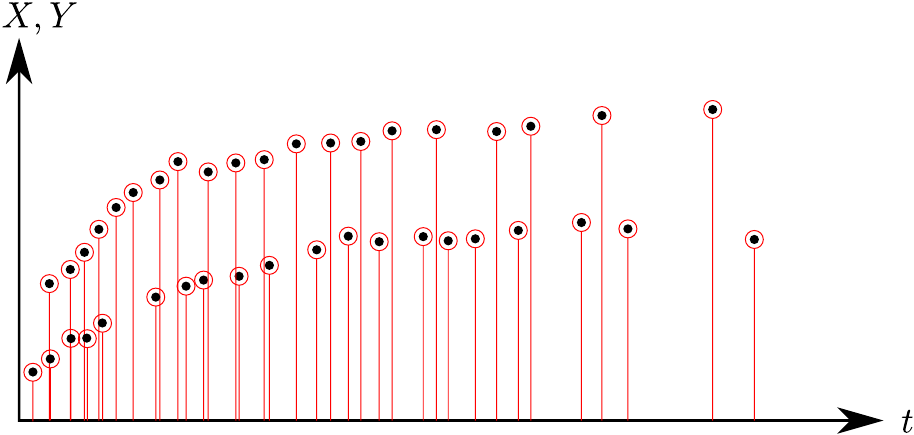}
 \caption{Event-driven optimisation}
 \label{fig:edopt}
 \end{subfigure} \quad
 \begin{subfigure}[b]{.3\linewidth}
 \centering
 \includegraphics[width=\textwidth]{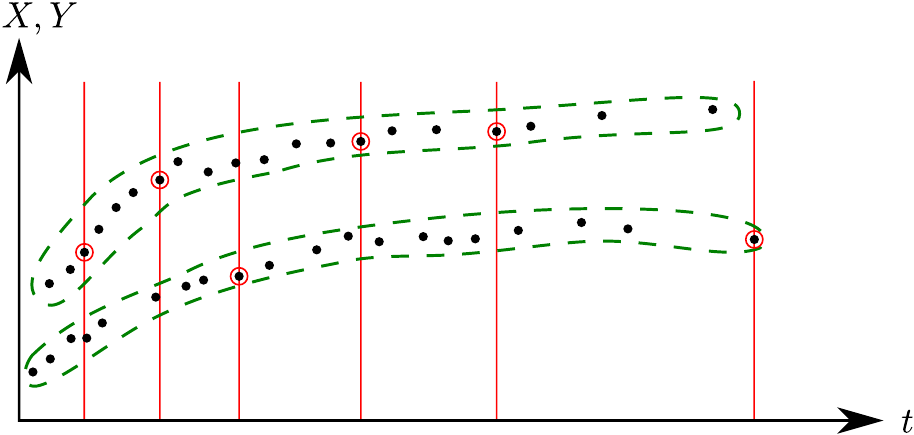}
 \caption{Event-driven and tracking opt.}
 \label{fig:edt}
 \end{subfigure}
 \caption{Contrasting batching and asynchronous processing for event-only VO. Black dots indicate an event stream unevenly distributed in the image plane $XY$ and time $t$. (a) Batching accumulates events in time windows, i.e., batches (represented by rectangular columns), to form event images $f_1, f_2, \ldots$, which are then subject to frame-based optimisation for VO. For correct operation, the batching parameters (e.g., location, duration and/or size of batches) must correctly reflect the underlying motion and scene structure. (b) Ideal event-driven optimisation integrates temporal and spatial information from each asynchronous event; the diagram stresses the event-driven sampling with a red decorator. (c) Event-driven optimisation with asynchronous event tracks (green dashed lines) increases efficiency by reducing optimisation instances (red lines).}
 \label{fig:backends}
\end{figure*}

A baseline approach is to temporally accumulate events to form event images (a.k.a.~\emph{batching}) then apply conventional frame-based VO methods, e.g.,~\cite{forster14, leutenegger15}. However, correct batching is critically important for frame-based methods. The optimal batching parameter (e.g., batch size or duration) is a time-varying quantity that depends on the instantaneous motion and scene complexity~\cite{Zhu_2017_CVPR,mueggler2015lifetime}. Arguably, finding the optimal batching parameter (in real-time) is a non-trivial problem itself~\cite{Zhu_2017_CVPR,mueggler2015lifetime}. Fig.~\ref{fig:fb} shows an inaccurate VO result by a frame-based method due to suboptimal batching. 


Fundamentally, batching of events leads to a loss of temporal and spatial information, thus lowering signal-to-noise ratio. An asynchronous approach for VO that 
\vspace{-2em}
\begin{itemize}
    \item (P1) Fits the trajectory using all \emph{individual} events $e_k$; and
    \item (P2) Updates the motion at each newly \emph{incoming} event
\end{itemize}
will help alleviate the problems due to batching. Fig.~\ref{fig:backends} conceptually contrasts batching and asynchronous processing.

A promising idea to realise asynchronous event-driven VO is to estimate a continuous-time (CT) camera trajectory from $\mathcal{E}$. This implicitly allows the camera pose to be temporally interpolated and extrapolated, which in turn provides a basis to perform P1 and P2. Along the lines above, there have been efforts on event-driven VO~\cite{mueggler15,mueggler18,wang21,gentil20}. However, previous works either assume a known map~\cite{mueggler15,mueggler18}, restrict the camera to planar motions~\cite{wang21}, or incorporate an IMU with preintegration~\cite{gentil20_preintegration} thereby achieving event-inertial VO~\cite{gentil20}.

Towards \emph{event-only} and \emph{map-free} monocular VO in $SE(3)$, we propose an asynchronous structure-from-motion (SfM) back-end optimisation that satisfies P1 and P2 above. Our formulation conducts incremental \emph{maximum a posteriori} (MAP) optimisation~\cite{dong17,dong18} to achieve CT trajectory estimation and mapping~\cite{barfoot14}. An $SE(3)$ trajectory is modelled using a Gaussian Process (GP), which allows the camera poses at all $e_1, e_2, \dots$ and the pertinent scene structure to be robustly reasoned from noisy event data. Moreover, the estimated CT trajectory can be extrapolated to allow motion updates at each new event. Fig.~\ref{fig:eovo} illustrates the benefits of our approach, while Sec.~\ref{sec:results} will present further empirical results. Our work also represents one of the first applications of~\cite{dong17,dong18,barfoot14} to event-based VO.

\section{Related work}

\subsection{Frame-based optimisation with batching}

Early event-based VO methods conduct batching to generate event images which are then fed into frame-based pipelines~\cite{rebecq16,rebecq17,zhu17,zhou18,zhou21}. Images are usually used to represent the event information for batching techniques, which can further be divided into two groups. The first group typically produces \emph{motion-compensated images}~\cite{rebecq16,rebecq17,gallego18,vidal18,liu20} by assuming some motion model~\cite{liu20} or relying on IMU measurements~\cite{rebecq16,rebecq17}, while the second group produces \emph{time-surface maps}~\cite{lagorce17,zhou18,zhou21} by the temporal component of the event. Most of the above methods use a fixed temporal resolution~\cite{zhu17} or spatial resolution~\cite{rebecq16, rebecq17,vidal18} to conduct batching, specified respectively by the time duration of a batch and the number of events in a batch.

Batching based on fixed temporal or spatial resolutions are a source of failure in frame-based VO. The example data stream in Fig.~\ref{fig:backends} corresponds to a decelerating camera in a static scene, which causes more events at the beginning of the stream and fewer events at the end. In this scenario, batching using a constant temporal resolution (the approach depicted in Fig.~\ref{fig:fbopt}) will eventually lead to images with insufficient information (e.g., no events in $f_5$) for motion estimation. On the other hand, batching based on constant spatial resolution in the decelerating camera scenario above will lead to noisy (``blurry'') event images, since a batch during the slow motion period is no longer a good representation of the instantaneous motion. Moreover, the optimal spatial resolution also depends on the scene complexity in the camera FOV~\cite{khan19}, which can change rapidly during the VO process.

In principle, a dynamic batching strategy~\cite{mohamed20} can alleviate the issues above. However, accurately estimating batching parameters is difficult; perhaps as difficult as performing VO itself since the optimal batching parameters depend on knowledge of the instantaneous camera motion. 




\subsection{Event-driven optimisation}

Event-driven optimisation for VO (see Fig.~\ref{fig:edopt}) should alleviate the weaknesses of batching. Fundamentally, however, a single event does not provide enough information to compute/update the camera motion, hence a motion model needs to be maintained by the algorithm~\cite{mueggler18} to facilitate asynchronous processing. Previous methods have mainly utilised non-parametric motion models~\cite{mueggler15,mueggler18,wang21,gentil20}.

Mueggler et al.~\cite{mueggler15} pioneered a CT framework for event-based VO with known map and later extended with IMU fusion~\cite{mueggler18}. Their method models the $SE(3)$ camera trajectory by using cubic splines. However, their approach will not work in unknown environments or when IMU is not available. The event-inertial approach of Le Gentil et al.~\cite{gentil20} jointly optimises scene features (3D lines in \cite{gentil20}) and the camera motions, based on GP motion model. However, their paradigm based on IMU preintegration~\cite{gentil20_preintegration} (regressing IMU pose and velocity from IMU readings) is fundamentally different from CT trajectory estimation. In~\cite{gentil20}, the role of GP is on preintegrating IMU and not on trajectory interpolation, and the resulting trajectory is a discrete set of camera poses and velocities. Closest in spirit to ours is~\cite{wang21} which is based on volumetric contrast maximisation to perform CT trajectory estimation (using splines) and jointly solving for camera tracking and (dense) mapping for event-only VO. However, only planar trajectories were considered.



As alluded to in Sec.~\ref{sec:intro}, our core contribution is to formulate an asynchronous back-end optimisation that can support map-free event-based VO in $SE(3)$.  Our back-end is based on incremental MAP optimisation for CT trajectory estimation, and mapping~\cite{dong17,dong18,barfoot14}, where a GP is formulated into CT $SE(3)$ trajectory and GTSAM~\cite{gtsam} is further incorporated for incremental optimisation. Key to our formulation is a tight coupling between GP trajectory modelling and MAP inference to solve data association and outlier (spurious event) removal. The incremental MAP estimation supports per-event updates (Fig.~\ref{fig:edopt}) as well as delayed updates (Fig.~\ref{fig:edt}) to improve efficiency if an event-driven feature tracker (e.g.,~\cite{alzugaray2020haste}) is available, without batching.


\begin{figure*}
    \centering
    \includegraphics[width=.9\linewidth]{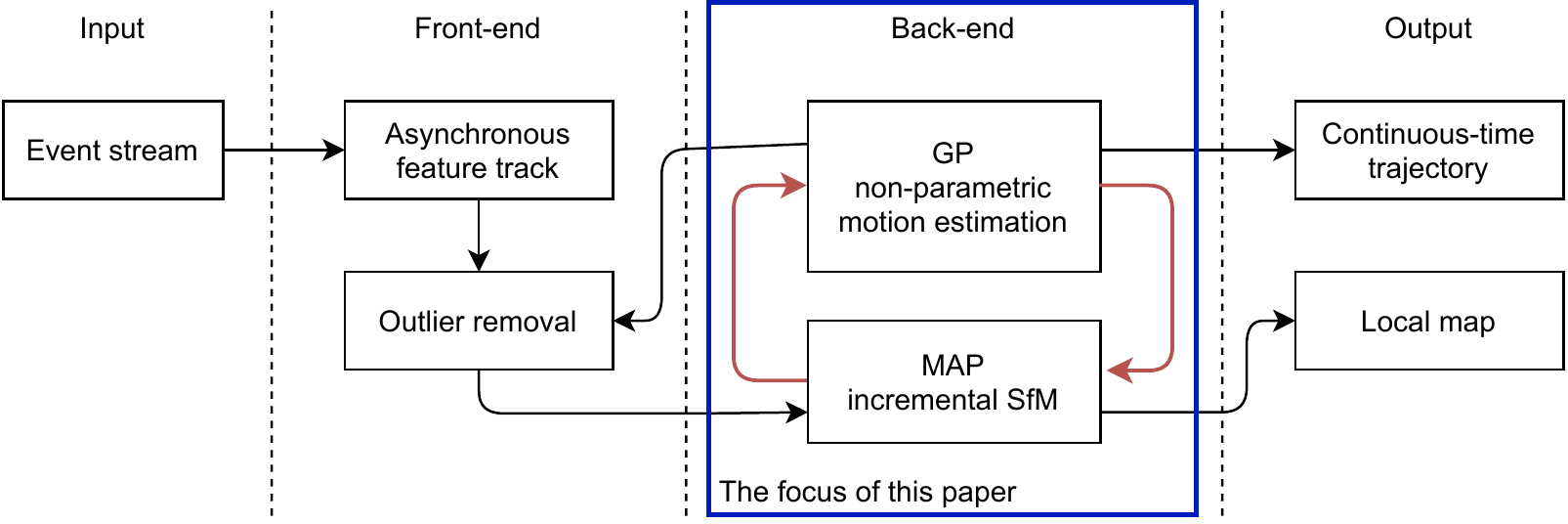} 
    \caption{Event-driven VO pipeline. The front-end receives a stream of asynchronous observations. The feature track component of the front-end solves data association. Then an outlier removal component filters out inconsistent measurements. The asynchronous back-end uses all spatial and temporal event information. The GP and MAP components collaborate to produce a continuous-time trajectory that can be updated from a single event. The back-end produces a local map as a sub-product. }
    \label{fig:pipeline}
\end{figure*}




\section{Event-driven VO pipeline}\label{sec:asyncvo}

Our event-driven VO pipeline is shown in Fig.~\ref{fig:pipeline}. The input to the pipeline is an event stream $\mathcal{E}$, and the output of the pipeline is a CT camera trajectory in $SE(3)$ and a local map which is a by-product. The camera trajectory is represented by a GP motion model. At the core of the pipeline---outlined by the blue box in the diagram, which is also the focus of our paper---is a MAP incremental SfM optimisation routine.

In the rest of the section, we will justify the design of the pipeline. Details of the core optimisation routine will be provided in Sec.~\ref{sec:formulation}, while some details of our implementation of the pipeline will be given in Sec.~\ref{sec:implemenetation}.



\subsection{GP representation for CT camera trajectory}

Let $q(t) \in SE(3)$ be a camera pose at time $t$. Our VO method employs the state representation
\begin{align}\label{eq:markovtraj}
    x(t) := (  q(t), \, \dot{q}(t) )
\end{align}
and models the event camera trajectory as a GP
\begin{align}\label{eq:post}
    x(t) \,\sim\, \GP(\mu(t),\cK(t,t'))
\end{align}
with mean $\mu(t)$ and covariance (kernel function) $\cK(t,t')$. We would like to fit~\eqref{eq:post} onto $\mathcal{E}$; in practice, this is equivalent to being able to compute the maximum of the posterior
\begin{align}\label{eq:regression}
    p(x(\tau) \,|\, \{x(t_i)\}_{i=1}^M, \, \mathcal{E} ),
\end{align} 
for any desired time $\tau$, where
\begin{align}
\{x(t_i)\}_{i=1}^M
\end{align}
is a set of $M$ \emph{control states}; these can be the states associated to a subset of the events (the red circles in Fig.~\ref{fig:edt}), or the states of all $\mathcal{E}$ (the scenario depicted in Fig.~\ref{fig:edopt}). Maximising~\eqref{eq:regression} is called \emph{GP regression}. Secs.~\ref{sec:formulation} and~\ref{sec:implemenetation} will describe GP regression and control state selection respectively.

Important characteristics of the GP representation are:

\paragraph{Compactness}

The appropriate number of control states depends on the complexity of the motion. In segments of the trajectory with uniform motions, the control states can be significantly fewer than event data. Arguably a discrete-time trajectory representation (e.g., camera poses defined over frames obtained from batching) is also more ``compact'' than the raw event data, however, it does not facilitate asynchronous processing, as explained next.


\paragraph{Asynchronous processing}

The ability to perform GP regression at any $\tau$ permits to seamlessly incorporate event data into the estimation of the control points, i.e., optimising over asynchronous measurements. On the other hand, a discrete-time trajectory representation, which does not allow smooth interpolation, will require a minimum amount of co-visibility (visual overlapping of the observations) between adjacent states to prevent a ``break'' in the trajectory.


\paragraph{Constant-time regression}

The benefits above arguably exist in other CT trajectory representations for event-based VO (e.g., splines~\cite{mueggler15,mueggler18}). However, a fundamental benefit of GP is constant-time regression, i.e., the cost of finding the maximum of~\eqref{eq:regression} is independent of the number of control states $M$. This is enabled by the Markovian structure of the state representation~\eqref{eq:markovtraj} (assuming a white-noise-on-acceleration prior~\cite{barfoot14}), which leads to a class of GPs with exactly sparse inverse kernel matrix~\cite{barfoot14}.





\subsection{Event-based MAP incremental SfM}

The goal of the back-end is to perform the computations to fit the GP model~\eqref{eq:post} onto the event stream $\mathcal{E}$ by estimating a set of control states. The problem translates to MAP inference of the control states from $\mathcal{E}$. Imperative for a VO system is to extend the trajectory in an online fashion, and to do so efficiently. For our GP formulation, this translates into incorporating new control states into the trajectory without solving the MAP problem from scratch. We call this problem MAP \emph{incremental SfM} (details in Sec.~\ref{sec:formulation}).

MAP incremental SfM has been explored previously, but \emph{not} for event-driven VO. To address planar trajectories from wheel and laser odometry, Yan et al.~\cite{yan15} presented an \emph{incremental} version of the approach in~\cite{barfoot14}. The key component for incremental MAP was using a Bayes tree~\cite{kaess10} (as available in GTSAM~\cite{gtsam}) to perform incremental variable reordering and just-in-time relinearisation operations during incremental MAP optimisation. Other works~\cite{dong17} have followed the same strategy: Anderson et al.~\cite{anderson15} solves GP trajectory regression in $SE(3)$ and Dong et al.~\cite{dong17} extended the approach to general matrix Lie groups. We based our $SE(3)$ implementation in the code of~\cite{dong17}.

\subsection{Asynchronous back-end integration}

Our MAP incremental SfM back-end enables the desired properties of an event-driven VO (P1 and P2 in Sec.~\ref{sec:intro}) to be achieved. There are also other appealing properties that support event-only VO pipeline (Fig.~\ref{fig:pipeline}):
\begin{itemize}
\item The estimated CT trajectory can facilitate tracking events in the front-end. For example, a GP trajectory extrapolation can guide local search for event data association. 
\item The outlier removal component can remove inconsistent observations assisted by GP constant-time interpolation.
\item Under the unified probabilistic framework of GP and MAP SfM, the back-end estimates the distribution of the trajectory. Thus, front-end components can introduce risk factors in their strategies.
\end{itemize}
\vspace{-1em}
\section{Continuous-time SfM formulation for event-only VO}\label{sec:formulation}

In this section, we provide the details of the core optimisation routine in the proposed event-drive VO pipeline (Fig.~\ref{fig:pipeline}).

\subsection{Measurements}

Let $\mathcal{E} = \{ e_k \}^{N}_{k=1}$ be the event data observed thus far. Let $z_k$ to be the pixel coordinates of an event at time $t_k$. Under a probabilistic Gaussian framework, 
\begin{align}\label{eq:projection}
z_k = h_k(x(t_k)) + \epsilon_k, \;\forall k=1,\ldots,N
\end{align}
where $h_k$ is the pinhole projection function (our measurement model) of the observed scene point and $\epsilon_k \sim \mathcal{N}(0, \mathcal{R}_k)$ is zero-mean Gaussian noise with covariance $\mathcal{R}_k$.

We group all observed events into the measurement vector
\begin{align}
Z := \left[\, z_1^T \; \cdots \; z_N^T \right]^T \,\sim\, \mathcal{N}(0,\mathcal{R})\,,
\end{align}
where $\mathcal{R}$ accommodates all covariances $\mathcal{R}_k$, $k=1,\ldots,N$.

\subsection{Variables}

Our GP formulation estimates a set of control states for the trajectory and a set of 3D landmarks. Define
\begin{align}
\mathcal{X} := \{ (x_i, s_i) \}_{i=1}^M 
\end{align}
as a set of $M$ control states $x_j := x(s_j)$ in correspondence with a subset of event timestamps
$\{ s_j \}_{j=1}^M \subseteq \{ t_k \}_{k=1}^N$
(usually $M \ll N$). Fundamentally, events correspond to brightness changes at 3D scene points; here, we assume that the control states have been associated to a set of 3D landmarks $\left[\ell_1^T \; \cdots \; \ell_L^T\right]^T$ (details of data association will be given in Sec.~\ref{sec:implemenetation}). Under GP modelling, the (unknown) landmarks are random variables
\begin{align}
\left[\ell_1^T \; \cdots \; \ell_L^T\right]^T \sim\mathcal{N}(\lambda,\mathcal{L})
\end{align}
drawn from a multivariate Gaussian. We group all variables in the vector
\begin{align}
X := \left[\,x_1^T \; \cdots \; x_M^T \; \ell_1^T \; \cdots \; \ell_L^T \,\right]^T \,\sim\, \mathcal{N}(\eta,\mathcal{P}),
\end{align}
where 
\begin{align}
\eta  := \left[\,\mu(s_1)^T \; \cdots \; \mu(s_M)^T \;\; \lambda^T \, \right]^T 
\end{align}
and
\begin{align}
\mathcal{P}  := 
\begin{bmatrix}
\cK & \\
& \mathcal{L}
\end{bmatrix}.
\end{align}
In the following, for simplicity we use the notation $x_i$ for the $i$-th element of $X$; the reader should take note that $x_i$ as an item of $X$ could refer to a landmark variable.


\subsection{MAP inference}\label{sec:map}

The Gaussian distribution $\mathcal{N}(\eta,\mathcal{P})$ defines the prior distribution in the MAP formulation. Our aim is find the $X$ that maximises the posterior density given the measurements $Z$:
\begin{subequations}
\begin{align}
\hat{X} &= \argmax_{X} p(X|Z)    \\
 &\propto \argmax_{X} p(Z|X) \, p(X). \label{eq:map}
\end{align}
\end{subequations}
Under Gaussian noise assumption, the likelihood 
\begin{align}\label{eq:likelihood}
     p(Z|X) \propto  \prod_k  \exp\left\{ -\dfrac{1}{2}  \| h_k(x(t_k)) - z_k \|_{\mathcal{R}_k}^2 \right\}
\end{align}
and the prior distribution 
\begin{align}\label{eq:prior}
     p(X) \propto \prod_{i} \exp\left\{ -\dfrac{1}{2}  \| x_i - \eta_i \|_{\cP_i}^2  \right\}
\end{align}
are well characterised. However, to conduct MAP optimisation, we need to write the likelihood~\eqref{eq:likelihood} in terms of the variables $\{x_i\}$. By using the GP representation for a Markovian trajectory, $x(t_k)$ can be written in terms of variables $x_l, \, x_r$, such that $(x_l, s_l), \, (x_r, s_r) \in \mathcal{X}$, and $s_l<t_k<s_r$ are the tightest variable timestamps to the event observation.  We stress that the likelihood~\eqref{eq:likelihood} incorporates all $N$ events into the MAP estimator~\eqref{eq:map} of the combined variables. 

\subsection{Trajectory Interpolation}\label{sec:trajectory_interpolation}


The GPs for Markovian trajectories resulting from LTV-SDE produce the $O(1)$ interpolation~\cite{barfoot14}
\begin{align}\label{eq:inter_vs}
    x(t_k) = \mu(t_k) + \Lambda(t_k)(x_l - \mu(s_l) ) + \Psi(t_k)(x_r-\mu(s_r)),
\end{align}
where 
\begin{align}
    \mu(t_k) := \Phi(t_k,s_l) \, \mu(s_l)
\end{align}
and the matrix operators $\Lambda$, $\Psi$ and $\Phi$ are as defined in~\cite{barfoot14}. As a linear combination of two trajectory states ($x_l, x_r$), GP interpolation is the key to efficiently introduce the observations into the CT SfM formulation. 

However, \eqref{eq:inter_vs} assumes trajectory states as elements of a vector space, which is invalid for trajectories in $SE(3)$. Dong et al.~\cite{dong17} addressed this limitation by extending this special class of GPs to matrix Lie groups~\cite{absil09}.  The state $x(t)$ of Markovian trajectory~\eqref{eq:markovtraj} in $SE(3)$, becomes
\begin{align}
\gamma(t; t_i) := (\xi(t),\, \dot{\xi}(t)),
\end{align}
in its Lie algebra $\mathfrak{se}(3)$, where $[\xi(t)]_{\times}$ is an element of $\mathfrak{se}(3)$ coinciding with the local tangent space of $SE(3)$. Precisely, for a transformation (a pose) $T(t) \in SE(3)$ around a linearisation point $T_i \in SE(3)$,
\begin{align}
[\xi(t)]_{\times} := T_i^{-1} T(t).
\end{align}
Thus, GP interpolation is also $O(1)$ in the tangent space
\begin{align}
\gamma(t_k; s_l) =  \Lambda(t_k) \gamma(s_l; s_l) + \Psi(t_k)\gamma(s_r; s_l) .
\end{align}
See~\cite{dong17,dong18} for more details.




\subsection{Solving the CT MAP problem}

Gauss-Newton (GN) method is typically used to solve MAP optimisation problems with non-linear measurement models. In the context of~\eqref{eq:map}, GN is applied to compute the locally optimal displacement $\delta X$ for the state update
\begin{align}\label{eq:update}
X + \delta X \rightarrow X 
\end{align}
by solving 
\begin{align}\label{eq:lls}
    \min_{\delta X} \dfrac{1}{2} \| J \delta X - \Delta Z  \|_{\mathcal{R}}^2 + \dfrac{1}{2} \| \delta X  - \Delta X  \|_{\mathcal{P}}^2,
\end{align}
where
\begin{align}
J := \dfrac{\partial h }{\partial X}
\end{align}
is the Jacobian matrix with $h:= [h_1^T, \cdots, h_N^T ]^T$, 
\begin{align}
\Delta X := X-\eta
\end{align}
is the \emph{deviation from the mean}, and 
\begin{align}
\Delta Z := Z-h
\end{align}
is the \emph{prediction error}. For brevity, we define the update~\eqref{eq:update} as a vector addition; however, this update is really carried out over a Lie manifold through the exponential map~\cite{absil09}. The least-squares problem~\eqref{eq:lls} has the normal form
\begin{align}\label{eq:ne}
(\mathcal{P}^{-1} +J^T \mathcal{R}^{-1} J ) \delta X^* = \mathcal{P}^{-1} \Delta X + J^T \mathcal{R}^{-1}  \Delta Z,
\end{align}
which allows $\delta X^*$ to be obtained using linear solvers.

\subsection{Sparsity}

Crucial to efficiently find the update $\delta X^*$ from~\eqref{eq:ne} is the sparsity of the \emph{information matrix} \begin{align}
A := \mathcal{P}^{-1} +J^T \mathcal{R}^{-1} J.
\end{align}
Since $A$ is symmetric, a recurrent strategy is to find the Cholesky factorisation $A = B^TB$ ($B$ is lower triangular) and then solve for $\delta X^*$ through back-substitution. The complexity of solving the Cholesky factorisation is affected by the sparsity of $A\in\mathbb{R}^{n\times n}$: from $O(n^3)$ for dense matrices to $O(n^1)$ for narrowly banded matrices~\cite{yan17}. 

The $J^T \mathcal{R}^{-1} J$ component in $A$ is the information from the (event) measurements. This matrix reflects the sparse structure of the SLAM problems~\cite{dellaert06}. 

\begin{align}
\mathcal{P}^{-1}  = 
\begin{bmatrix}
\cK^{-1} & \\
& \mathcal{L}^-1
\end{bmatrix}
\end{align}
encodes the prior information. It is also sparse as $\cK^{-1}$ is sparse for Markovian trajectories~\cite{barfoot14} and $\mathcal{L}$ is diagonal.









\section{Implementation}\label{sec:implemenetation}

Algorithm~\ref{alg:imp} presents the main steps of our implementation of the event-driven VO pipeline introduced in Sec.~\ref{sec:asyncvo}. An important problem is solving for data association. Our implementation uses the asynchronous method of~\cite{alzugaray2020haste} (Lines \ref{step:track_start} and  \ref{step:track_cont}) to track event features (Lines \ref{step:feat_start} and \ref{step:feat_cont}) obtained from a simple heuristic based on histogrammming events.



\begin{algorithm}
\begin{algorithmic}[1]
    \STATE Set a list of control states $\mathcal{X} = \{\}$.
    
    \STATE Listen for an initial event $e_0 = (z_0, t_0, p_0)$.
    \STATE Add a new control state $(x_0, t_0)$ to  $\mathcal{X}$, where $x_0$ encodes the initial pose and velocity at time $t_0$.
    
    \STATE Listen for more events to track the selected features. \label{step:feat_start}
    
    \STATE {Track the selected features; for example, with~\cite{alzugaray2020haste}.} \label{step:track_start}
    
    \WHILE{Tracking}
    \IF {there is only one state in $\mathcal{X}$}
    \STATE {Add $(x_1, t_c)$ to $\mathcal{X}$ such that $x_1$ encodes the rel. pose between matching keypoints at $t_0$ and current $t_c$. } \label{step:pose}
    \STATE{Estimate the initial map coordinates by triangulating matching keypoints.}
    \STATE {Refine the estimates by solving the CT SfM to reason over all events within the event tracks.} 

    \ENDIF
    
    \STATE Add a new $(x_{\text{new}}, t_c)$ to $\mathcal{X}$ from GP extrapolation at current time $t_c$.
    
    \STATE Remove outlying tracks as those with average reprojection error (over all its events) above some threshold. 
    
    \STATE Optimise variables with our asynchronous back-end.
    
    \IF {number of event tracks is less than a threshold}
        \STATE Find more feature tracks as in Line~\ref{step:feat_start}. \label{step:feat_cont}
    \ENDIF

    \STATE {Continue tracking the selected features.} \label{step:track_cont}

    \ENDWHILE
    
    

\end{algorithmic}
\caption{Asynchronous VO}
\label{alg:imp}
\end{algorithm}

\section{Results}\label{sec:results}

\subsection{Evaluation of core VO optimisation}

\begin{figure} [h]
\centering
 \begin{subfigure}[b]{\linewidth}
 \centering
 \includegraphics[width=0.85\textwidth]{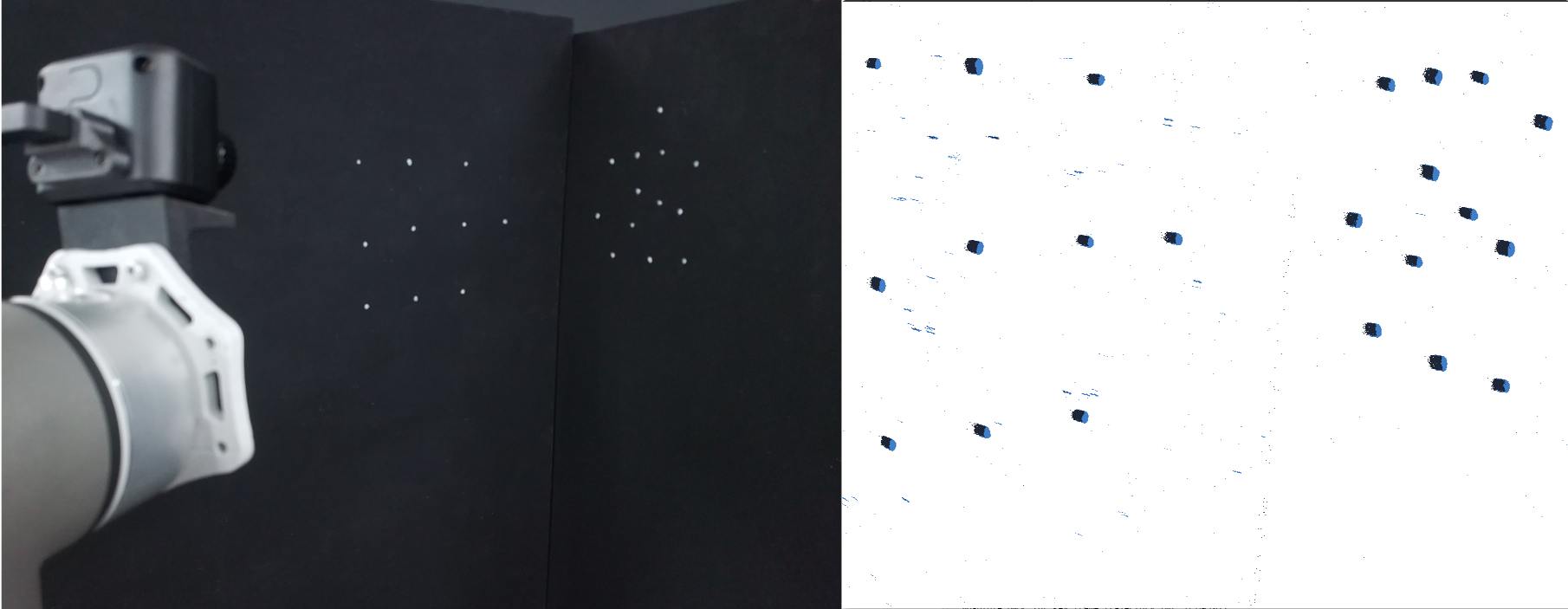}
 \caption{Recording of event data and sample event image.}
 \label{fig:set_up}
 \end{subfigure} 
 \begin{subfigure}[b]{\linewidth}
 \centering
 \includegraphics[width=0.85\textwidth]{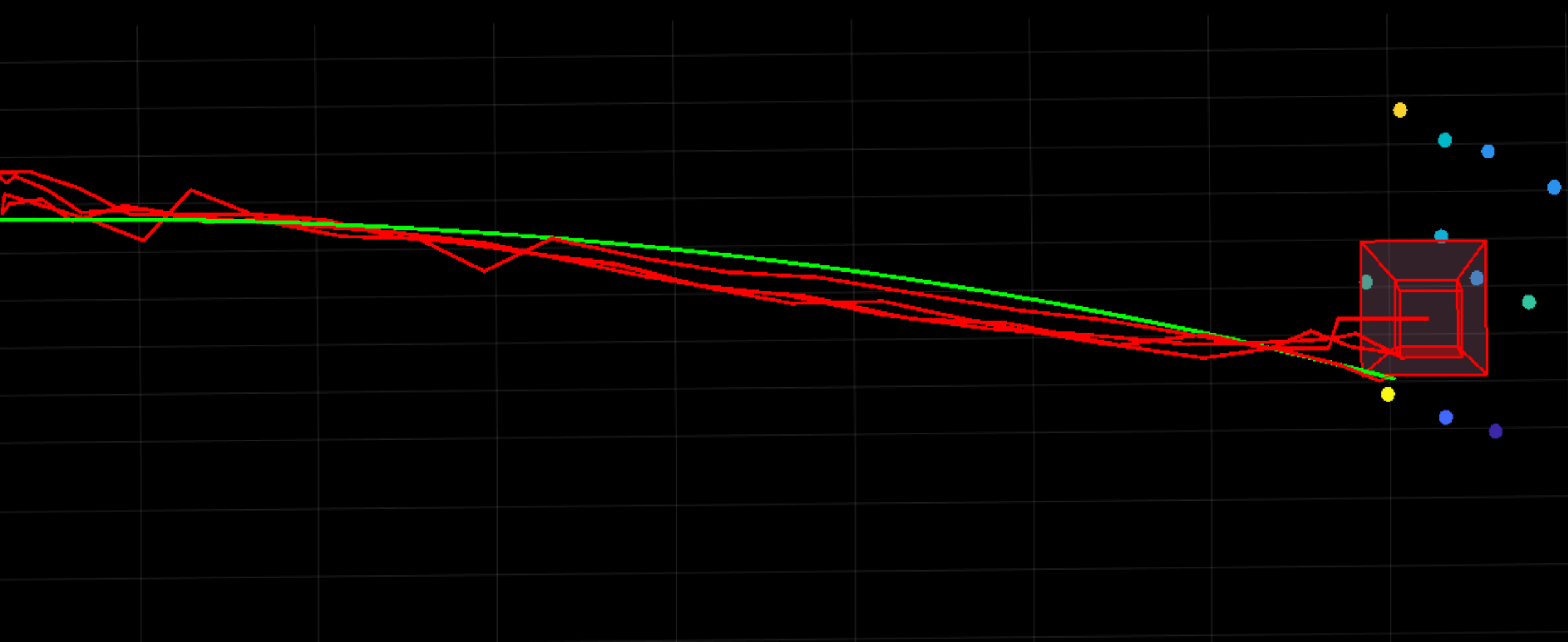}
 \caption{Frame-based optimisation (red: estimated, green: ground truth).}
 \label{fig:seq01_bun}
 \end{subfigure}
 \begin{subfigure}[b]{\linewidth}
 \centering
 \includegraphics[width=0.85\textwidth]{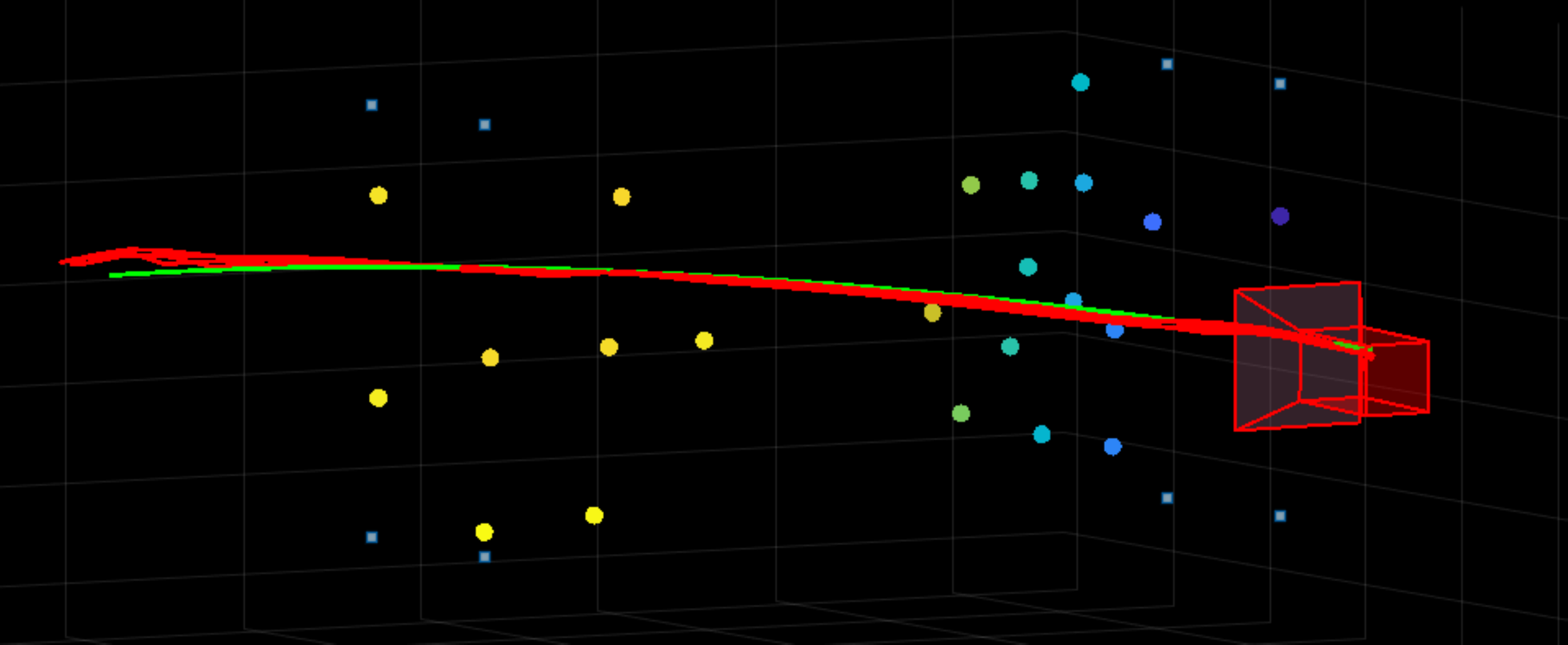}
 \caption{Asynchronous optimisation (red: estimated, green: ground truth).}
 \label{fig:seq01_GP}
 \end{subfigure}
 \caption{(a) Recording of event data with a Prophesee camera undergoing repetitive motion with a robot arm. (b)(c) Estimated trajectories from the frame-based back-end (bundle adjustment) and asynchronous back-end (MAP incremental SfM with GP model). Dots are the estimated scene points.}
 \label{fig:seq01_result}
\end{figure}

First, we focus on evaluating the core optimisation routine (Sec.~\ref{sec:formulation}) using data captured under controlled settings. We recorded a 30-second event sequence (containing $\approx$10M events) from a simple scene with several dots with a Prophesee event camera mounted in an UR5 robot arm; see Fig.~\ref{fig:set_up} for the setup. The simple scene enabled the usage of STR~\cite{liu2021spatiotemporal} (essentially point cloud registration) to conduct data association. The preprocessed data is then subjected to the MAP incremental SfM routine. For comparisons, we generated event images with 100 ms batches, then subjected the frames (with data association performed by associating the average coordinates of feature locations in frames) to bundle adjustment. Qualitative and quantitative results are depicted in Fig.~\ref{fig:seq01_result} and Table~\ref{tab:seq1}; the latter is by computing the relative pose error (RPE) and absolute trajectory error (ATE) (see~\cite{sturm2012benchmark} for the definitions) between the estimated trajectory and ground truth trajectory of the robot arm. The results show that, if data association is satisfactory, both asynchronous and frame-based back-ends perform equally well; of course, the latter also presumes accurate batching, which may not be possible in all cases (Sec.~\ref{sec:intro}). In addition, the smoother trajectory in the asynchronous back-end is due the GP prior. Using our unoptimised implementation of MAP incremental SfM, the result in Fig.~\ref{fig:seq01_GP} was generated in $\approx$1 minute (excluding preprocessing).


\begin{table}[]
\begin{center}
\begin{tabular}{l|cc}
\hline
Method             & Asynchronous & Bundle Adjustment \\ \hline
RPE [m] &   \textbf{0.0124}               &  0.0136                 \\
ATE [m] &   \textbf{0.0954}              &    0.1066     \\
\hline
\end{tabular}
\caption{Relative Pose Error (RPE) and Absolute Trajectory Error (ATE) (see~\cite{sturm2012benchmark} for the definitions) for our asynchronous backend and frame-based bundle adjustment.}
\label{tab:seq1}
\end{center}
\vspace{-4em}
\end{table}

\subsection{Evaluation of VO pipeline}

\begin{figure} [h]\centering
 \begin{subfigure}[b]{0.45\linewidth}
 \includegraphics[width=\textwidth]{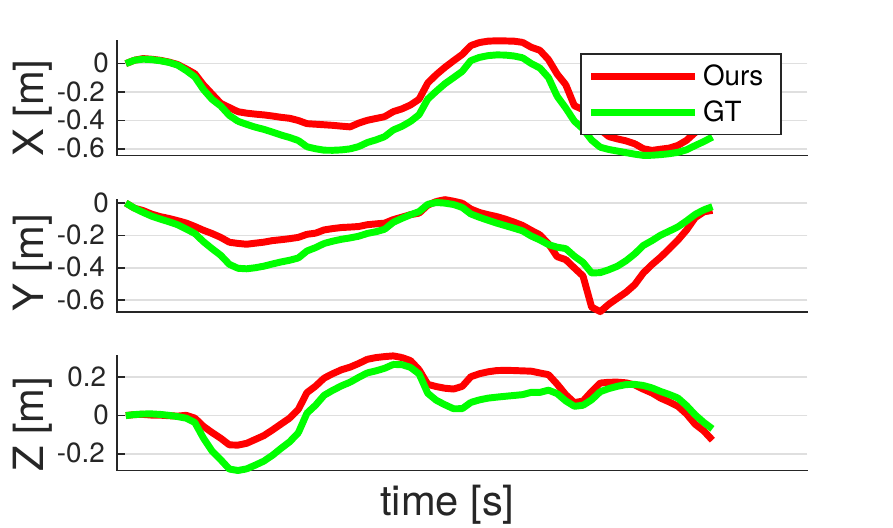}
 \caption{}
 \label{fig:xyz_dynamic1}
 \end{subfigure} 
 \begin{subfigure}[b]{0.45\linewidth}
 \includegraphics[width=\textwidth]{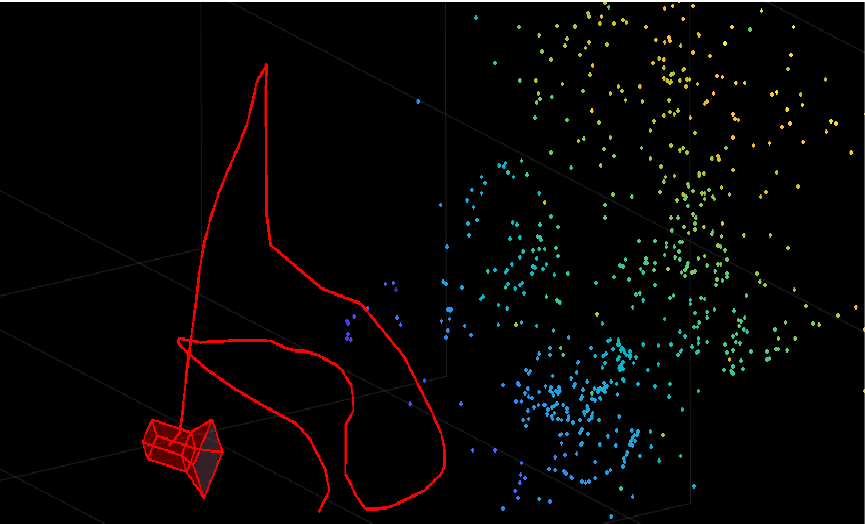}
 \caption{}
 \end{subfigure} 
 \begin{subfigure}[b]{0.45\linewidth}
 \includegraphics[width=\textwidth]{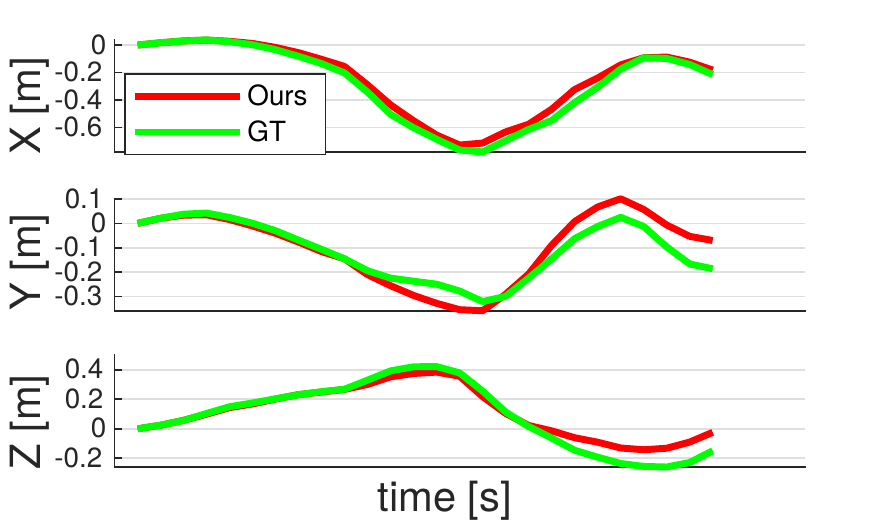}
 \caption{}
 \label{fig:xyz_dynamic2}
 \end{subfigure} 
 \begin{subfigure}[b]{0.45\linewidth}
 \includegraphics[width=\textwidth]{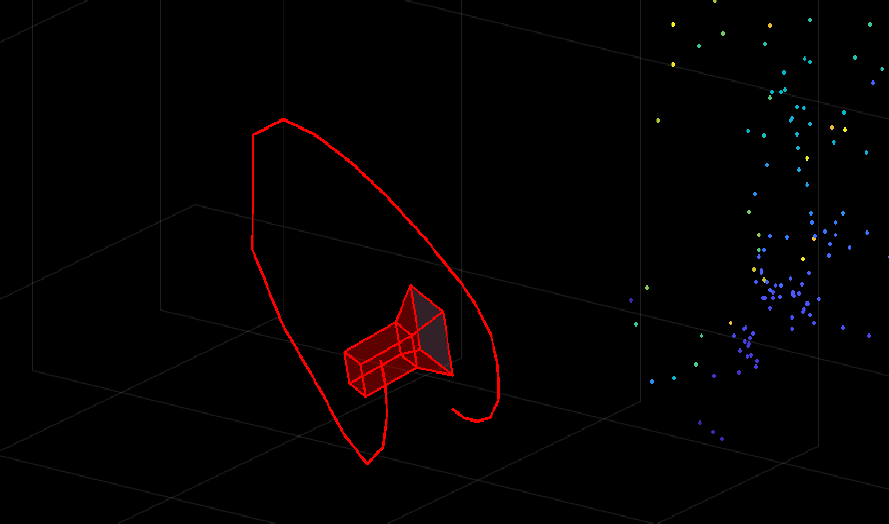}
 \caption{}
 \label{fig:trajectory_dynamic2}
 \end{subfigure} 
 \begin{subfigure}[b]{0.45\linewidth}
 \centering
 \includegraphics[width=\textwidth]{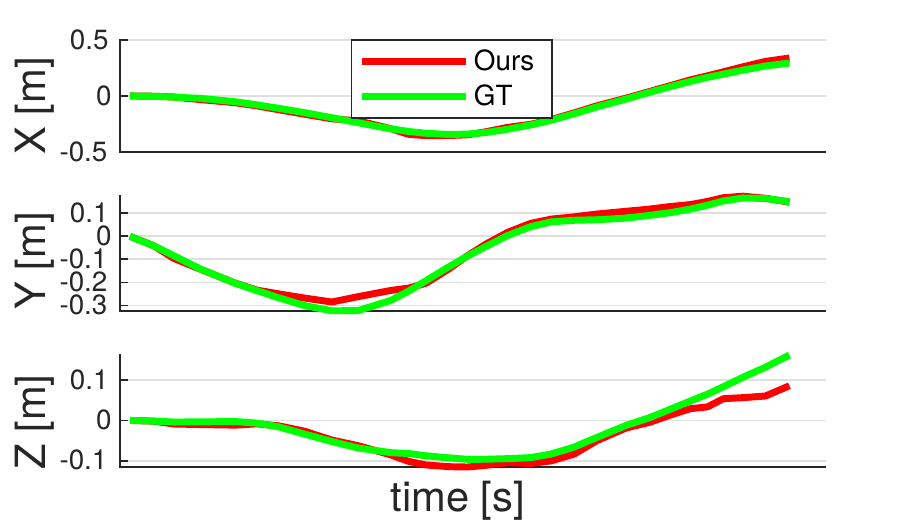}
 \caption{}
 \label{fig:xyz_dynamic3}
 \end{subfigure} 
 \begin{subfigure}[b]{0.45\linewidth}
 \includegraphics[width=\textwidth]{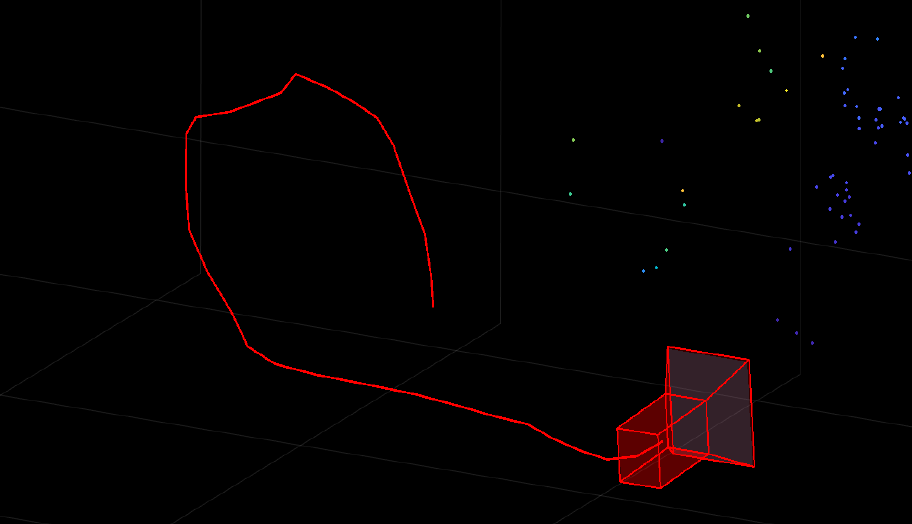}
 \caption{}
 \label{fig:trajectory_dynamic3}
 \end{subfigure} 
 \caption{Event-based VO results from three subsequences (rows 1 to 3) of \texttt{Dynamic\_6dof}. (a)(c)(e) Coordinates $(X,Y,Z)$ of the trajectory estimated by Algo.~\ref{alg:imp} compared against ground truth. (b)(d)(f) Estimated camera trajectory.}
 \label{fig:dynamic}
 \end{figure}

To evaluate Algorithm~\ref{alg:imp}, we employed publicly available event data, specifically the \verb|Dynamic_6dof| sequence from the Event-Camera Dataset~\cite{mueggler2017event} (see~\cite{mueggler2017event} for the detailed description of the dataset). At this stage, our implementation of Algorithm~\ref{alg:imp} is rudimentary, thus our program is not able to perform VO continuously throughout the sequence. However, our program is able to work successfully on subsequences of 5 s-duration ($\approx$5 M events in each sequence) from \verb|Dynamic_6dof|; see Fig.~\ref{fig:dynamic}. Currently, our implementation is also not optimised; the total runtime of the back-end was $\approx$10 minutes to execute on each subsequence. Nonetheless, we believe our results indicate the promise of our asynchronous back-end for event-only VO.

\vspace{-1em}

\section{Conclusions and future work}
We presented an event-driven back-end for event-only VO which integrates event observations in a principled way with a continuous-time SFM formulation. We believe our proposed back-end provides theoretical justification on how to achieve a practical event-only VO system. To this goal, we believe significant effort should be devoted to designing ad hoc mechanisms for front and back-end collaboration that leverage a CT trajectory representation. This integration could benefit typical sub-processes required in VO pipelines such as new point creation, data association and outlier removal. 

\section*{Acknowledgement}
This work was supported by Australian Research Council ARC DP200101675.
\vfill





\vfill

\pagebreak

\bibliographystyle{IEEEtran}
\bibliography{async}

\end{document}